%% file: arxiv_main.tex
\documentclass[10pt,twocolumn,letterpaper]{article}
\input{macros}

\iccvfinalcopy % *** Uncomment this line for the final submission

\begin{document}

%%%%%%%%% TITLE
\title{Dynamic Point Fields}

\author{
Sergey Prokudin\affmark[1]\quad
Qianli Ma\affmark[1,2]\quad
Maxime Raafat\affmark[1] \quad
Julien Valentin\affmark[3] \quad
Siyu Tang\affmark[1]\quad
\\
\affaddr{\affmark[1]ETH Z\"urich}\quad
\affaddr{\affmark[2]Max Planck Institute for Intelligent Systems} \quad  \affaddr{\affmark[3]Microsoft}\\
% \email{firstname.lastname@inf.ethz.ch}
}

\maketitle

%%%%%%%%% ABSTRACT

\begin{abstract}
    Recent years have witnessed significant progress in the field of neural surface reconstruction. While the extensive focus was put on volumetric and implicit approaches, a number of works have shown that explicit graphics primitives such as point clouds can significantly reduce computational complexity, without sacrificing the reconstructed surface quality. However, less emphasis has been put on modeling dynamic surfaces with point primitives. In this work, we present a dynamic point field model that combines the representational benefits of explicit point-based graphics with implicit deformation networks to allow efficient modeling of non-rigid 3D surfaces. Using explicit surface primitives also allows us to easily incorporate well-established constraints such as-isometric-as-possible regularisation. While learning this deformation model is prone to local optima when trained in a fully unsupervised manner, we propose to also leverage semantic information such as keypoint dynamics to guide the deformation learning. We demonstrate our model with an example application of creating an expressive animatable human avatar from a collection of 3D scans. Here, previous methods mostly rely on variants of the linear blend skinning paradigm, which fundamentally limits the expressivity of such models when dealing with complex cloth appearances such as long skirts. We show the advantages of our dynamic point field framework in terms of its representational power, learning efficiency, and robustness to out-of-distribution novel poses. The code will be made publicly available \footnote{\label{footnote_supp}\url{https://sergeyprokudin.github.io/dpf}}.
        
\end{abstract}

\section{Introduction}
\label{sec:intro}

\begin{figure}[t]
\begin{center}
  \includegraphics[trim={0.5cm 0.0cm 0.0cm 0cm},clip,width=1.0\linewidth]{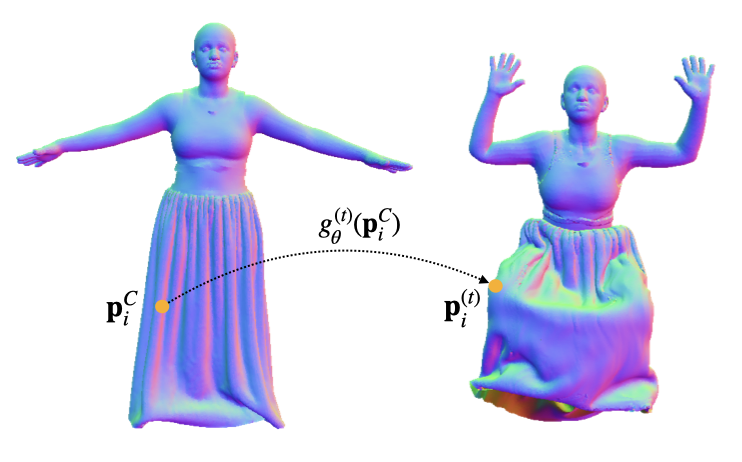}
\end{center}

  \caption{\textit{Dynamic point field}. We propose to model dynamic surfaces with a point-based model, where the motion of a point $\mathbf{p}_i$ over time is represented by an implicit deformation field $g^{(t)}_\theta$. Working directly with points rather than SDFs allows us to easily incorporate various well-known deformation constraints, e.g. as-isometric-as-possible \cite{kilian2007geometric}. We showcase the usefulness of this approach for creating animatable avatars in complex clothing.}
  \vspace{-0.3cm}
\label{fig:fig1}
\end{figure}

\begin{figure*}[t!]
\centering
\includegraphics[trim={0.0cm 0.0cm 0.0cm 0.cm},clip,width=\textwidth]{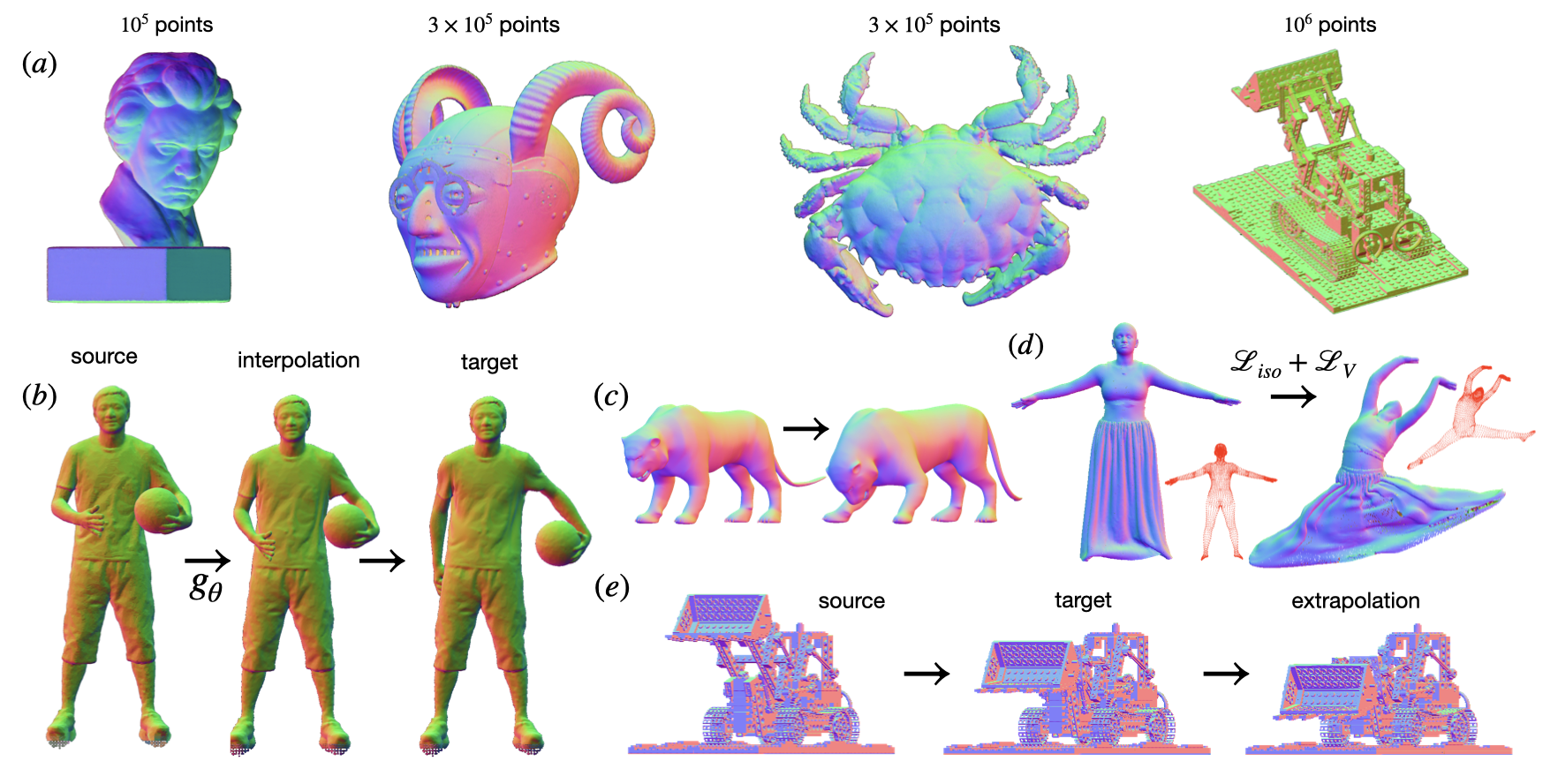}
\vspace{-0.2cm}
\caption{\textit{Overview of the main results.} \textit{(a)}: static surface reconstruction with optimised point clouds. We propose a point cloud optimisation scheme that efficiently utilises available off-the-shelf point renderers \cite{wiles2020synsin,pytorch3d} and surpasses the performance of more sophisticated methods \cite{takikawa2021nglod,mueller2022instant}, with an explicit surface representation which requires zero inference time. \textit{(b-e)}: we extend the approach to allow modeling dynamic surfaces of varying topology and complexity, and propose a guided learning for the case of complex deformations. Our formulation also allows us to interpolate easily between canonical and target surfaces.}
\label{fig:fig2}
\end{figure*}

Neural surface reconstruction and rendering have been subject to giant leaps in the last couple of years \cite{tewari2020state}. While original works have suffered from the representational limitations stemming from using a single global neural network to represent a surface as implicit signed distance \cite{Atzmon_2020_CVPR,Park_2019_CVPR,sitzmann2019siren} or occupancy field \cite{mescheder2019occupancy}, recent models pushed the level or reconstruction quality by introducing auxiliary data structures such as octrees \cite{takikawa2021nglod} or hashmaps \cite{mueller2022instant}. While providing high level of accuracy, these lines of work rely on the ability to perform large amounts of queries to the underlying multi-layer perceptrons (MLPs) to reconstruct the surface. 

In the field of volumetric neural rendering, this restriction has motivated development of compact and computationally attractive models that can train quickly and render at interactive rates \cite{garbin2021fastnerf, reiser2021kilonerf,yu2021plenoctrees,Chen2022ECCV,liu2022devrf}.  Among them, point-based methods for neural rendering \cite{lassner2021pulsar,wiles2020synsin,aliev2020neural,xu2022point,ruckert2022adop,Rakhimov_2022_CVPR,zhang2022differentiable} proved to be scalable alternatives to purely implicit or volumetric approaches. In the context of differentiable neural surface modeling, points have also been explored for static scenes \cite{peng2021shape,Yifan:DSS:2019} and dynamic human modeling \cite{POP:ICCV:2021}.

This work aims to achieve three goals. First, we showcase the advantage of a point-based surface representation compared to the most recent implicit models \cite{mueller2022instant}. We do this by carefully analyzing the behavior of the existing framework when representing complex 3D meshes with perfect ground truth available. Here, point clouds as primitives demonstrate direct benefits over their implicit alternatives in all the comparison dimensions: reconstructed geometry quality, training time, runtime, and model size.

Second, we extend the point-based surface model to support {\it non-rigidly deforming surfaces}. A number of approaches has been proposed to tackle this problem, both for generic 3D scenes \cite{innmann2020nrmvs,novotny2022keytr} and human-specific models \cite{POP:ICCV:2021,zakharkin2021point}. However, these models often tend to constrain the space of deformations via restrictions to specific algebraic operations, commonly linear blend skinning functions \cite{POP:ICCV:2021} or a mixture of affine warps \cite{Lombardi:2019}. Instead, we rely on a deformation field represented by a neural network \cite{palafox2021npms}, similar to the approach taken in dynamic neural radiance fields \cite{pumarola2020d,park2021nerfies}. Again, we show that learning such deformation networks on point sets is easier and more robust compared to SDFs. 
More importantly, it also allows us to easily incorporate various well-known regularisation techniques, such as as-isometric-as-possible \cite{kilian2007geometric,huang2008non}, enforcing the preservation of distances between points in the canonical and deformed space.

Further, we propose a guided learning regime to improve the robustness of the dynamic point fields and allow 3D shape manipulation with our framework. Our key idea is the following: undesirable local optima can be avoided by exploiting sparse correspondences in space-time as additional constraints, further improving optimisation speed and stability.
For instance, to reconstruct the highly dynamic and complex deformation of a long skirt (Figure \ref{fig:fig5}), we supervise the learning of the deformation field with the body vertices correspondences between frames from the readily available underlying unclothed body model registrations \cite{POP:ICCV:2021}. As demonstrated in the experiments section, the correspondence constraints on the minimally-clothed bodies provide strong generalization capabilities on the deformation of a long skirt largely deviating from the body surface. Based on this observation, we propose a method for zero-shot avatar reposing based on the introduced deformation framework. Our method compares favorably to other point-based \cite{POP:ICCV:2021,SkiRT:3DV:2022} and implicit methods
\cite{saito2021scanimate} which are fundamentally based on the linear blend skinning paradigm and fail to correctly represent challenging cloth types.

To summarise, our main contributions are as follows. We first introduce \textit{dynamic point fields}, a simple and computationally attractive model combining the compactness and efficiency of point primitives with the flexibility and accuracy of neural networks to model deformations. Advantages of the proposed approach are demonstrated by comparing it with the state-of-the-art in various surface reconstruction and deformation learning tasks. Second, we propose a \textit{guided deformation field learning} and show how clever constraints on available keypoint correspondences can be used to efficiently guide the learning of surface dynamics, without sacrificing the expressivity of a model.

\section{Related Work}
\label{sec:related}

\paragraph{Point-based representations.} Point clouds have a long history of applications in computer vision, both for scene rendering and surface modeling \cite{gross2011point,kobbelt2004survey,levoy1985use,pfister2000surfels,rusinkiewicz2000qsplat}. With the rise of differentiable rendering techniques \cite{tewari2020state}, they gained a renowned interest and in many use cases, have shown to be efficient alternatives to both volumetric and mesh-based techniques \cite{wiles2020synsin,Yifan:DSS:2019,aliev2020neural,lassner2021pulsar,ruckert2022adop,xu2022point, kopanas2022neural}. Our paper mainly uses the differentiable point rendering technique proposed in \cite{wiles2020synsin} and \cite{lassner2021pulsar} for point cloud optimisation, which is in turn similar in spirit to differential surface splatting \cite{Yifan:DSS:2019}.

\paragraph{Differentiable surface representations.} The early age of visual computing was mostly dominated by explicit representations such as meshes and volumes. However, recent advances have shown the benefits of implicit representations, where a signed distance field or an occupancy is parameterised via a deep neural network \cite{Park_2019_CVPR,mescheder2019occupancy}. Follow-up works have further improved expressivity and efficiency of the deep representations by changing the network structure and activations \cite{sitzmann2019siren,Peng2020ECCV} or incorporating regularisations \cite{Atzmon_2020_CVPR,icml2020_2086}. 
Further augmenting neural networks with auxiliary data structures such as octrees \cite{takikawa2021nglod} or hashmaps \cite{mueller2022instant} has also demonstrated to significantly boost model performance. Compared to prior works, these models have shown the capacity to represent highly detailed surfaces with many intricate details. We mostly focus on comparing to these baselines, as they represent the state-of-the-art in the surface modeling field. Similar to~\cite{Peng2021SAP}, we show that a simple point cloud, which naturally handles 3D space sparsity, can efficiently rival the modern aforementioned approaches, while offering numerous benefits such as model simplicity, explicitness and compactness.

\paragraph{Dynamic surface modeling, non-rigid registration.} Modeling deformations has in turn a long history in computer vision \cite{miller2006geodesic,von2006vector,luthi2017gaussian,ovsjanikov2012functional,puy2020flot,feydy2019interpolating,hirose2020bayesian,myronenko2010point,amberg2007optimal,sumner2007embedded}. Among the most intensively studied are the models that are built on top of 3D templates for the learning of dynamic shapes \cite{Dyna:SIGGRAPH:2015,alldieck2018video,huang2017towards,groueix20183d,corona2021smplicit,loper2015smpl,SMPL-X:2019}. Several point-based models have recently been introduced to facilitate neural rendering and surface modeling of dynamic humans \cite{POP:ICCV:2021,prokudin2021smplpix,SkiRT:3DV:2022}. However, these models are still limited when it comes to modeling challenging clothes and dynamics such as skirts, due to heavy reliance on linear blend skinning functions.
In contrast, our model is comprised of a simple and expressive point set and a flexible neural network to model non-rigid deformations of varying complexity.  The formulation is also more general than recently proposed point-based methods for human draping \cite{zakharkin2021point}.
Modeling generic point cloud deformations has been addressed in several works \cite{wand2007reconstruction,li2018efficient,stoll2006template,shi2021graph}. However, many works escape local minima by restricting the set of possible deformations \cite{innmann2020nrmvs} or building priors on the cloud motion \cite{myronenko2010point,eisenberger2018divergence}.

In contrast, we utilise recent developments in implicit dynamic scene and surface modeling \cite{niemeyer2019occupancy,park2021hypernerf,park2021nerfies,li2022neural,tiwari21neuralgif,lei2022cadex}. Similar to \cite{pumarola2020d}, we employ a deep neural network that learns to transfer points from a canonical frame to a target frame. Our differentiating factor here is the use of point clouds as our canonical space graphics primitive instead of SDFs or radiance fields. Second, in the case of challenging human surface modeling, using explicit 3D primitives also enables the direct supervision of our deformation network with motion of minimally clothed body model, which significantly boosts the training time and helps avoiding local optima.

\paragraph{Clothed human modeling and animation.}

Given a set of 3D clothed human scans, how do we animate them into new poses with visually plausible clothing shapes? 
A majority of recent clothed human models address this problem by learning a regression model from the body pose to the clothing geometry, using a variety of shape representations such as meshes~\cite{CAPE:CVPR:20,patel20tailornet,gundogdu2019garnet}, implicit surfaces~\cite{tiwari21neuralgif,saito2021scanimate,wang2021metaavatar,chen2021snarf} and point sets~\cite{zakharkin2021point,POP:ICCV:2021,SkiRT:3DV:2022,lin2022fite}. These methods typically rely on Linear Blend Skinning (LBS) to handle body articulation. While efficient, using LBS limits the models' capability in representing loose clothing and certain common garment types such as skirts and dresses. 
We refer the reader to \cite{SkiRT:3DV:2022} for a detailed discussion on the limitation of LBS-based clothed human models.
In this paper, we demonstrate that our guided deformation learning serves as an effective alternative in animating clothed humans. Being free from the drawbacks of LBS, it can generate coherent and plausible clothing shape even under extreme unseen poses.

\section{Method}
\label{sec:method}

\subsection{Dynamic Point Field Model}
\label{subsec:dpf}

Our framework comprises two elementary components: a point cloud with learnable spatial locations and features, and a set of compact neural networks (one for each time step) that warp every point into a new location. 

\paragraph{Dynamic point set.} More formally, we define a point cloud $X$ as a set of tuples $\mathbf{p}_i$: 
\vspace{-2pt}
\begin{eqnarray}
\mathbf{X} = \{ \mathbf{p}_i = (\mathbf{x}_i, \mathbf{n}_i), \mathbf{x}_i \in \mathbb{R}^3, \mathbf{n}_i \in \mathbb{R}^{3} \}_{i=1,\dots,N_p},
\label{eq:point_set}
\end{eqnarray}
where $\mathbf{x}_i$ are the 3D point locations and $\mathbf{n}_i$ the corresponding point normals. The point cloud consists of $N_p$ points.

We then represent a dynamic scene $\mathbf{S}$ as sequences of point sets:  
\vspace{-2pt}
\begin{eqnarray}
\mathbf{S} = [\mathbf{X}^{(0)}, \dots, \mathbf{X}^{(t)},   \dots, \mathbf{X}^{(T)}],
\label{eq:dynamic_scene}
\end{eqnarray}
where $\mathbf{X}^{(t)}$ is the point cloud representing the scene at time step $t$, and $T$ is the total number of steps.

\paragraph{Deformation field.} We aim to model dynamics of a scene~(Eq.~\ref{eq:dynamic_scene}) via a set of compact neural networks $\{g^{(t)}_{\theta}, t=1,\dots,T\}$, which update locations of every point  $\mathbf{p}_i^{\mathcal{C}}$ from a canonical set $\mathbf{X}^{\mathcal{C}}$: 

\vspace{-20pt}
\begin{center}
\begin{eqnarray}
g^{(t)}_{\theta}: \mathbb{R}^{3} \rightarrow \mathbb{R}^{3}, \\
\mathbf{x}_i^{(t)}= \mathbf{x}_i^{\mathcal{C}} +g^{(t)}_{\theta}(\mathbf{x}_i^{\mathcal{C}}).
\label{eq:deformation_network}
\end{eqnarray}
\end{center}

The corresponding warped normals $\mathbf{n}_i^{(t)}$ can be estimated in a fully differentiable manner using several approaches discussed in the appendix. In general, thanks to the isometric loss described in the next section and a bias of the network towards smooth predictions, the predicted deformations in our experiments allow us to directly transfer and reuse the canonical space mesh connectivity for estimation. 

Applying the deformation network to a set of points is simply equal to applying it to every point in the set:
\vspace{-2pt}
\begin{eqnarray}
\mathbf{X}^{(t)} =  g^{(t)}_{\theta}(\mathbf{X}^{\mathcal{C}}) = \{ \mathbf{x}_i^{\mathcal{C}} + g^{(t)}_{\theta}(\mathbf{x}_i^{\mathcal{C}})\}_{i=1,\dots,N_p}.
\label{eq:deformation_network_set}
\end{eqnarray}

 We set $\mathbf{X}^{\mathcal{C}} = \mathbf{X}^{(0)}$, if not stated otherwise. $\theta$ are the parameters of the neural networks.

\paragraph{Network architecture.} We use a multi-layer perceptron with periodic activations as our deformation network for its performance and flexibility in representing various input domains \cite{sitzmann2019siren}. If not stated otherwise, we use a small network with three hidden layers, each of size 128. This compact architecture makes running the deformation module on large point clouds extremely efficient. As an avenue for future work, we also consider using a meta-learning approach for learning these deformation networks based on a time-dependent latent variable, similar to  \cite{sitzmann2020metasdf,wang2021metaavatar}. In this work, we focus on a simple scenario where every target deformation is modeled with a dedicated small module.

We will now discuss the ways to efficiently optimize dynamic point fields.

\subsection{Training}
\label{subsec:training}

\paragraph{Canonical surface reconstruction.} Given a ground truth mesh $\mathcal{M}_{gt}$, for each optimisation step we obtain a point cloud  $\mathbf{X}_{gt}$ by sampling points from a mesh with the associated normal directions. We then optimise a combination of the following losses:
\vspace{-2pt}
\begin{eqnarray}
\mathcal{L}_{S}(\mathbf{X}, \mathcal{M}_{gt})= \lambda_{CD}\mathcal{L}_{CD} + \lambda_{n}\mathcal{L}_{n}+ \lambda_{n_I}\mathcal{L}_{n_I},
\label{eq:surface_loss}
\end{eqnarray}
\noindent where $\mathcal{L}_{CD}$ and $\mathcal{L}_{n}$ are the standard Chamfer-based discrepancy losses between ground truth and optimised clouds and corresponding normals \cite{pytorch3d}. $\mathcal{L}_{n_I}$ is the image-space normal discrepancy between rendered mesh normal image $\mathcal{I}^{n}_{gt}$ and point normal renders $\mathcal{I}^n_{X}$:

\vspace{-2pt}
\begin{eqnarray}
\mathcal{L}_{n_I}(\mathbf{X}, \mathcal{M}_{gt}) = ||\mathcal{I}^n_{gt}  - \mathcal{I}^n_{X}||_2.
\label{eq:image_normal_loss}
\end{eqnarray}

Here, images $\mathcal{I}^n_{gt}$ and $\mathcal{I}^n_{X}$ are obtained by rendering ground truth mesh and optimised point cloud normals from the same random camera position $C$ sampled from a unit sphere:
\vspace{-2pt}
\begin{eqnarray}
\mathcal{I}^n_{gt} = \mathcal{R}_{m}(\mathcal{M}_{gt}, C), \\
\mathcal{I}^n_{X} = \mathcal{R}_{p}(\mathbf{X}, C), 
\label{eq:normal_renderers}
\end{eqnarray}
\noindent where $\mathcal{R}_{m}$ is a mesh renderer, and $\mathcal{R}_{p}$ is the differentiable point-based renderer \cite{wiles2020synsin} available as a part of \cite{pytorch3d}. Compared to solely optimizing for Chamfer-based normal loss, enforcing image-based normals produces more visually appealing and consistent point normals. Please see the appendix for the loss analysis. 

We use $\lambda_{CD}=10^4, \lambda_{n}=1$ and $\lambda_{n_I}=10^1$ for balancing our loss functions and bringing them to the same scale.

\paragraph{Dynamic surface reconstruction.} Given a sequence of ground truth meshes
\vspace{-2pt}
\begin{eqnarray}
\mathbf{M}^{\sim}_{gt} = [\mathcal{M}^{(0)}_{gt}, \dots, \mathcal{M}^{(t)}_{gt},   \dots, \mathcal{M}^{(T)}_{gt}],
\label{eq:dynamic_meshes}
\end{eqnarray}
the above described formulation allows direct extension for the case of dynamic surface reconstruction with a deformation network:
\vspace{-2pt}
\begin{eqnarray}
\mathbf{X}^{\mathcal{C}*}, \theta^* = \argmin_{\mathbf{X}^{\mathcal{C}}, \theta}\sum_{t=1}^T{\mathcal{L}_{S}\big(\mathbf{X}^{(t)}, \mathcal{M}^{(t)}_{gt}}\big), \\
\mathbf{X}^{(t)} = g^{(t)}_{\theta}(\mathbf{X}^{\mathcal{C}}).
\label{eq:dynamic_surface_loss}
\end{eqnarray}

\begin{figure*}[t!]
\centering
\includegraphics[trim={0.0cm 0.0cm 0.0cm 0.0cm},clip,width=\textwidth]{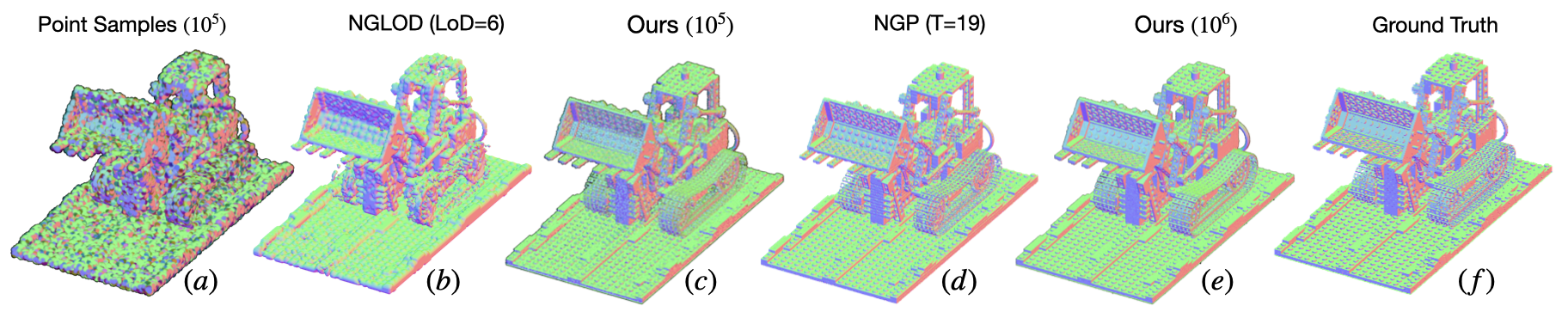}
\vspace{-0.4cm}
\caption{\textit{Representing 3D surfaces with optimisable point sets.} Compared to the state-of-the-art implicit models for 3D surface representation \cite{mueller2022instant,takikawa2021nglod}, optimised point cloud model offers better reconstruction quality on all metrics, while taking zero inference time thanks to its explicit nature. See Section~\ref{subsec:static_experiments} for details.} 
\label{fig:fig3}
\end{figure*}

However, the optimisation fails to find a plausible deformation field in the case of rapid and complex motion. To alleviate the problem, we propose two techniques for learning plausible deformations.

\paragraph{As-isometric-as-possible constraint.} Our explicit 3D surface formulation allows us to use some classical constraints for learning deformations in 3D space. In this work, we use the as-isometric-as-possible constraint to guide our learning \cite{kilian2007geometric}, which enforces the preservation of distances between points in the canonical and deformed space:

\vspace{-2pt}

\begin{eqnarray}
\mathcal{L}_{iso}= \sum_{i=1}^{N_p} \sum_{j \in \mathcal{N}_k(i)} ||d(\xb_i^\mathcal{C}, \xb_j^\mathcal{C}) -  d(\xb_i^{(t)}, \xb_j^{(t)})||_1,
\label{eq:iso_loss}
\end{eqnarray}
where $\mathcal{N}_k(i)$ is a $k$-neighborhood of the point $\xb_i$ in $\mathbf{X}^\mathcal{C}$. We set $k=5$, if not stated otherwise.

To promote meaningful interpolations (see Figure \ref{fig:fig2}\textit{b}), we can also enforce the same constraint on the intermediate deformations:

\vspace{-10pt}
\begin{eqnarray}
\mathcal{L}^{\gamma}_{iso}= \sum_{i=1}^{N_p} \sum_{j \in \mathcal{N}_k(i)} ||d(\xb_i^\mathcal{C}, \xb_j^\mathcal{C}) -  d(\xb_i^{(\gamma t)}, \xb_j^{(\gamma t)})||_1, \\
\mathbf{x}_i^{(\gamma t)}= \mathbf{x}_i^{\mathcal{C}} + \gamma \cdot g^{(t)}_{\theta}(\xb_i^{\mathcal{C}}),
\label{eq:iso_loss_inter}
\end{eqnarray}
where $\gamma \in [0,1]$. Enforcing this loss for $\gamma \in [1, \inf)$ allows us to also generate plausible extrapolations outside of the motion interval. See Figure \ref{fig:fig2}\textit{e} for an example of extrapolation of excavator bucket motion.

\paragraph{Guided deformation field learning.} In many scenarios, paired information on the transformation of certain 3D points and their features is available:

\begin{eqnarray}
\mathbf{V}^{(t)} = \{(\mathbf{v}_i^\mathcal{C}, \mathbf{v}_i^{(t)})\}_{i=1,\dots, N^{(t)}_v},
\label{eq:keypoint_supervision}
\end{eqnarray}
where $(\mathbf{v}_i^\mathcal{C}, \mathbf{v}_i^{(t)}) $ are the locations and features of a $i^{th}$ keypoint in the canonical and target frames respectively, and $N^{(t)}_v$ is the total number of available keypoints per frame $t$.  We denote the full set of keypoints for a scene~(Eq.~\ref{eq:dynamic_scene}) as:
\vspace{-2pt}
\begin{eqnarray}
\mathbf{V}_{S} = [\mathbf{V}^{(0)}, \dots, \mathbf{V}^{(t)},   \dots, \mathbf{V}^{(T)}].
\label{eq:scene_keypoints}
\end{eqnarray}

This information can come from either 3D feature matching between canonical and target frames \cite{lepard2021}, 3D keypoints \cite{openpose}, or via registration or regression of a certain parametric model to 3D surfaces \cite{loper2015smpl}. In the latter case, $(\mathbf{v}_i^\mathcal{C}, \mathbf{v}_i^{(t)})$ are simply 3D model mesh vertices in the canonical and posed spaces. We can utilise this information as a direct supervision signal for our deformation field model:
\vspace{-2pt}
\begin{eqnarray}
\mathcal{L}_{V} = \sum_{i=1}^{N^{(t)}_v} ||\mathbf{v}_i^{(t)} - \big(\mathbf{v}_i^\mathcal{C} + g^{(t)}_\theta(\mathbf{v}_i^\mathcal{C})\big)||_1.
\label{eq:keypoint_loss}
\end{eqnarray}

We then combine it with our surface loss and isometric losses to form the following final loss:
\vspace{-2pt}
\begin{eqnarray}
\mathcal{L}(\mathbf{X}^\mathcal{C}, \theta, \mathbf{V}_S, \mathbf{M}^{\sim}_{gt}) = \lambda_{S}\mathcal{L}_{S} + \lambda_{iso}\mathcal{L}_{iso}  + \lambda_{V}\mathcal{L}_{V}.
\label{eq:full_dynamic_loss}
\end{eqnarray}

Regularly, we set $\lambda_{S}=0, \lambda_{V}=1, \lambda_{iso}=0.1$ in the early epochs to allow accelerated learning of large-scale deformations, and then gradually increase the weight of a surface loss to allow high-frequency deformation modeling. 

\paragraph{Avatar animation.} Setting $\lambda_{S}=0$ also allows us to perform a single scan animation, as we will demonstrate in Section \ref{subsec:avatars}. In this setup, the optimisation will deform a mesh to the desired target pose while respecting the introduced geometric constraints, similar in spirit to classic algorithms in shape manipulation \cite{kilian2007geometric,sorkine2007rigid,sorkine2004laplacian}.

\section{Experiments}
\label{sec:experiments}

\vspace{-20pt}

\begin{center}
\begin{table}
\footnotesize
\centering
\resizebox{\columnwidth}{!}{
\begin{tabular}{l|c|c|c|c}
Method                                                                & $\mathcal{L}_{CD}$ $\downarrow$ & $\mathcal{L}_{n}$ $\downarrow$  & Size (Mb) & Params ($10^6$)  \\
\hline 
  DSS \cite{Yifan:DSS:2019} ($10^5$ points)                                    & 3.060                 & 0.909                & 2.4    & 0.6                        \\
  SAP \cite{Peng2021SAP} ($10^4$)  & 0.856               & 0.764                 & 0.24   & 0.06                       \\                        
  SAP \cite{Peng2021SAP} ($10^5$)   & 0.659	 & 0.678                 &  2.4    & 0.6                          \\
  SAP \cite{Peng2021SAP} ($10^6$)   &      0.536	& 0.640        & 24      & 6   \\
   NGLOD \cite{takikawa2021nglod} (LoD=4)                                                         & 2.862                & 0.755                & 5.61   & 1.35                       \\
   NGLOD \cite{takikawa2021nglod}(LoD=5)                                                         & 1.832                & 0.748                & 38.7    & 10.14                      \\
  NGLOD \cite{takikawa2021nglod} (LoD=6)                                                         & 1.776                & 0.696                & 300     & 78.18                      \\
  NGP \cite{mueller2022instant} (T=11)                                                    & 3.429                & 0.887                & 0.29    & 0.07                       \\
  NGP \cite{mueller2022instant}(T=16)                      & 0.422                & 0.681                & 7.21    & 1.8                        \\
  NGP \cite{mueller2022instant} (T=19)                                                    & 0.361                & 0.468                & 48.8    & 12.2                       \\
  Ours ($10^4$ points)     & 3.362                & 0.626                & 0.24    & 0.06                       \\
  Ours ($10^5$) & 0.765                & 0.468                & 2.4     & 0.6                        \\
  Ours ($3\cdot10^5$) & 0.419                & 0.411                & 7.2     & 1.8                        \\
  Ours ($10^6$) & \textbf{0.246}       & \textbf{0.322}       & 24      & 6        \\ \hline 

\end{tabular}
}
\vspace{0.8pt}
  \caption{\textit{Static surface reconstruction with point sets.} Results reported on the Lego excavator 3D scene. $\mathcal{L}_{CD}$: Chamfer distance ($\times 10^{-4}$). $\mathcal{L}_{n}$: normal consistency. See Section \ref{subsec:static_experiments} for details. }
  \label{table:static}
\end{table}
\end{center}

\vspace{-10pt}
\subsection{Canonical Surface Reconstruction} 
\label{subsec:static_experiments}

We compare our optimized clouds to the two state-of-the-art implicit methods for static surface overfitting: neural geometric level of detail (NGLOD \cite{takikawa2021nglod}) and instant neural graphics primitives (NGP \cite{mueller2022instant}). Contrary to earlier works \cite{Park_2019_CVPR,sitzmann2019siren,sitzmann2019srns,niemeyer2019occupancy,icml2020_2086}, these methods are able to represent extremely complex surfaces, and hence they appear in the focus of the present work. For NGLOD, we explore the quality-memory trade-off by varying the level of detail parameter of the method (LOD). For the NGP method, same can be achieved by varying the size $T$ of a hash table that stores spatial features.  Our explicit point-based method clearly outperforms both implicit methods, while using a similar or smaller number of parameters. This motivates our usage of points for representing canonical space when modeling deformations. We also compare with two alternative point optimisation techniques \cite{Yifan:DSS:2019,Peng2021SAP}, and provide a discussion on the difference between approaches in the appendix. The shapes-as-points technique \cite{Peng2021SAP} can be complementary to our basic pipeline in the case of reconstruction from unoriented, noisy point clouds.

\begin{figure*}[t!]
\centering
\includegraphics[trim={0.0cm 0.0cm 0.0cm 0.0cm},clip,width=\textwidth]{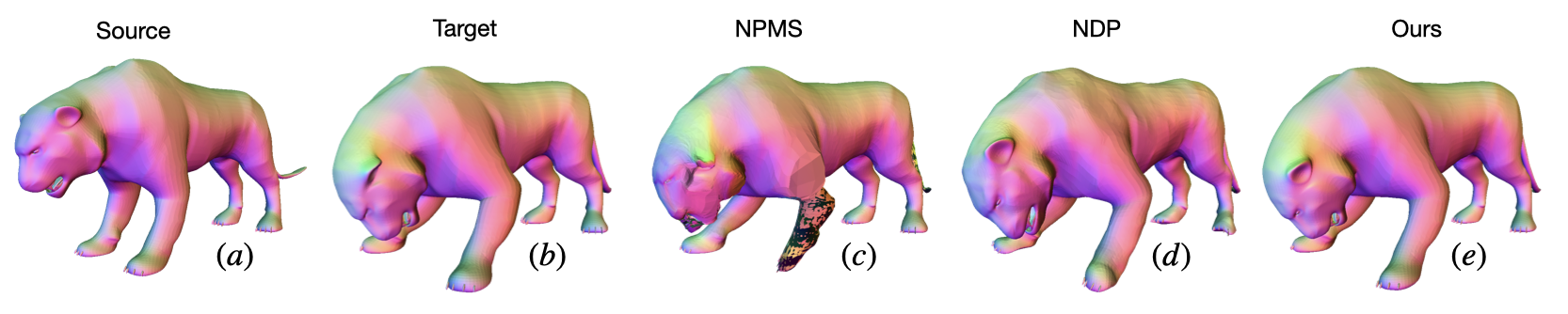}
\vspace{-0.5cm}
\caption{\textit{Unsupervised learning of non-rigid deformations.} We compare our deformation model to both SDF-based model \cite{palafox2021npms} and state-of-the-art non-rigid point cloud registration techniques \cite{li2022non} on diverse types of surfaces, and show qualitative and quantitative advantages of our model, both in terms of reconstruction and registration quality. Please pay attention to the reconstruction of the paws and ear locations of the presented feline model. See Section~\ref{subsec:deformation_fields_sdf} for details.} 
\label{fig:fig4}
\end{figure*}

\begin{table*}[t]
  \resizebox{\textwidth}{!}{
\begin{tabular}{l|ccccc|ccccc|cc|cc}
             & \multicolumn{5}{c|}{DeformingThings4D \cite{li20214dcomplete}}                                                                                                                    & \multicolumn{5}{c|}{Lego \cite{mildenhall2020nerf}}                                                                                                             & \multicolumn{2}{c|}{Owlii  \cite{xu2017owlii}}                       & \multicolumn{2}{c}{ReSynth  \cite{POP:ICCV:2021} }                    \\  \hline
             
             Method        &  $\mathcal{L}_{CD}$ $\downarrow$     & $\mathcal{L}_{n}$ $\downarrow$ & EPE $\downarrow$ & $Acc_S$ $\uparrow$  & $Acc_R$ $\uparrow$  & $\mathcal{L}_{CD}$  $\downarrow$  & $\mathcal{L}_{n}$ $\downarrow$ & EPE $\downarrow$ & $Acc_S$ $\uparrow$ &  $Acc_R$ $\uparrow$ &$\mathcal{L}_{CD}$ $\downarrow$ & $\mathcal{L}_{n}$ $\downarrow$ & $\mathcal{L}_{CD}$ $\downarrow$ & $\mathcal{L}_{n}$ $\downarrow$ \\  \hline
1 NPMs \cite{palafox2021npms}       &  198.8	& 0.305	& 0.233 & 27.2 &	34.6	& 9.351	& 0.432	& 0.012	& 91.1	& 93.0 &	2.168 &	0.072 &	312.2 &	0.401             \\
2 NSFP   \cite{li2021neural}             & 16.3                          & 0.38                    & 0.162                   & 30.6                       & 47.6                       & 0.644                  & 0.823                   & 0.021                   & 68.3                       & 90.1                       & 6.789                  & 0.309                   & 5.33                   & 0.449                   \\
3 Nerfies \cite{park2021nerfies}      & 12.1                          & 0.351                   & 0.211                   & 41.7                       & 53.2                       & 1.458                  & 0.883                   & 0.039                   & 37.2                       & 71.6                       & 1.631                  & 0.305                   & 3.345                  & 0.409                   \\
4 NDP \cite{li2022non}                 & 2.819 & 0.178                   & 0.122                   & 39.4                       & 63.1                       & 0.759                  & 0.896                   & 0.021                   & 71.1                       & 97.8                       & 1.365                  & 0.221                   & 3.889                  & 0.372                   \\
5 Ours (w/o $\mathcal{L}_{iso}$) & 1.045                         & 0.11                    & 0.222                   & 27.2                       & 34.4                       & 0.376                  & 0.713                   & 0.008                   & 96.1                       & \textbf{99.2}              & \textbf{0.111}         & \textbf{0.068}          & 0.613                  & 0.112                   \\
6 Ours (w. $\mathcal{L}_{iso}$)  & \textbf{0.786}                & \textbf{0.076}          & \textbf{0.103}          & \textbf{59.8}              & \textbf{69.1}              & \textbf{0.166}         & \textbf{0.416}          & \textbf{0.003}          & \textbf{97.1}              & 98.3                       & 0.123                  & 0.079                   & \textbf{0.309}         & \textbf{0.096}          \\
\hline
\multicolumn{15}{l}{\textit{Supervised w. \cite{lepard2021}:}}                                                                                                                                                                                                                                                                                                                                    \\ \hline  
7 LNDP \cite{li2022non}             & 3.769                         & 0.192                   & 0.115                   & 46.7                       & 69.1                       & 1.051                  & 0.896                   & 0.026                   & 54.2                       & 94.2                       & 3.559                  & 0.27                    & 9.959                  & 0.489                   \\
8 Ours (w. $\mathcal{L}_{iso}$)  & \textbf{0.644}                & \textbf{0.066}          & \textbf{0.099}          & \textbf{72.6}              & \textbf{80.4}              & \textbf{0.204}         & \textbf{0.495}          & \textbf{0.004}          & \textbf{96.8}              & \textbf{99.3}              & \textbf{0.116}         & \textbf{0.081}          & 0.188         & 0.097    \\ \hline  
\multicolumn{15}{l}{\textit{Supervised w. SMPL-X:}}   
\\ \hline  
9 NPMs \cite{palafox2021npms}  &  -                & -          & -          & -              & -              & -         & -          & -          & -              & -              & -         & -          & 7.258         & 0.164    \\
10 Ours (w. $\mathcal{L}_{iso}$)  &  -                & -          & -          & -              & -              & -         & -          & -          & -              & -              & -         & -          & \textbf{0.165}         & \textbf{0.071}    \\

\hline  
\end{tabular}
}
\vspace{0.8pt}
  \caption{\textit{Dynamic surface modeling with point fields.}  See Section \ref{subsec:deformation_fields_sdf} for the detailed discussion.}
  \label{table:dynamic}
  % \vspace{-10pt}
\end{table*}

\subsection{Learning Deformation Fields} 
\label{subsec:deformation_fields_sdf}
Our deformation model is based on a neural network that maps every point in the canonical space to a point in the deformed space. The idea was introduced before in the context of deformable radiance fields \cite{park2021nerfies,pumarola2020d}. For implicit signed distance fields, a similar deformation framework has been introduced in the context of deformable human modeling \cite{palafox2021npms}. In this section, we will show that learning deformation networks directly on explicit graphics primitives bring a number of benefits, including faster training, higher reconstruction fidelity, and lower memory requirements. Another additional benefit of working with explicit surfaces is the ability to directly enforce well-known constraints, such as as-isometric-as-possible deformation.

\paragraph{Datasets and metrics.} To achieve this goal, we select target and source deformation surfaces from several datasets of varying complexity: 10 pairs for the DeformingThings4D dataset (Figure \ref{fig:fig2}\textit{c}), 5 pairs from the Owlii dynamic scans \cite{xu2017owlii} (Figure \ref{fig:fig2}\textit{b}), 5 pairs from the synthetic Resynth dataset \cite{POP:ICCV:2021} and a single source-target pair of a Lego scene (Figure \ref{fig:fig2}\textit{e}). These datasets imply different combinations of surface complexity and non-rigidity at different scales (skeletal motion, high-frequency cloth wrinkles), making it a hard task for deformation learning. We use two types of metrics to compare various deformation architectures and supervision regimes. First, for all datasets, we measure  how well the deformation can match the target shape. We use the same Chamfer distance and normal metrics for this as in the previous section. Second, for the DeformingThings4D dataset and Lego, we additionally measure the point registration quality by comparing the endpoint error (EPE) and strict and relaxed 3D accuracies $Acc_S$, $Acc_R$, metrics utilized in the current state-of-the-art approach for non-rigid point cloud registration \cite{li2022non}.

\paragraph{Baselines.} For the SDF baseline, we consider the SDF deformation framework introduced in \cite{palafox2021npms}. The main focus of this section is to compare the deformation learning process on SDFs and explicit primitives. We, therefore, are leaving aside the latent shape and pose space models also introduced by the approach, and simply compare the capability to learn a single non-trivial deformation of a surface. To simplify the deformation learning procedure, we also use pre-trained best-performing SDF fields for target and source shapes, obtained with an instant NGP framework. In this way, the experiment can fully focus on simply learning the deformation field between two shapes, rather than simultaneously addressing the reconstruction of canonical shape and field deformation. 

We also compare to the recent state-of-the-art non-rigid shape registration method (NDP \cite{li2022non}), which has been shown to outperform a large number of previously introduced baselines \cite{ovsjanikov2012functional,puy2020flot,feydy2019interpolating,hirose2020bayesian,myronenko2010point,amberg2007optimal}. The method is generally similar in spirit to our proposed point deformation network, varying  on the utilized network architectures and deformation constraints. While this approach has been shown to perform robustly on the data with relatively smooth and small deformations \cite{lepard2021}, it is interesting to consider its applicability in the case of rapid body motion and high-frequency surface deformations. We have also experimented with the SE-3 variant of deformation parametrization introduced in \cite{park2021nerfies}, as well as neural scene flow prior approach \cite{li2021neural}. 

\begin{figure}
\begin{center}
  \includegraphics[trim={0.0cm 0.0cm 0.0cm 0cm},clip,width=1.0\linewidth]{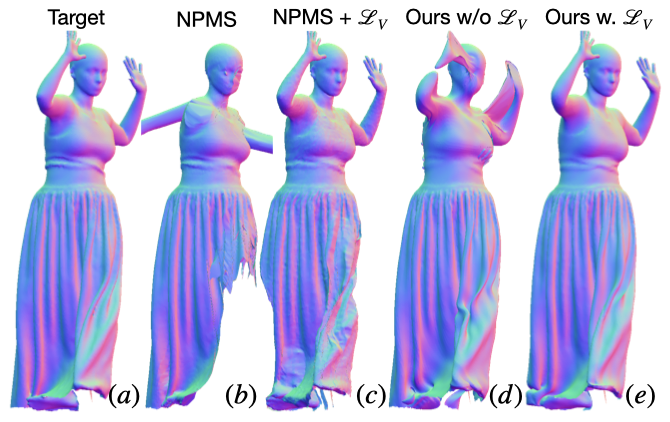}
\end{center}
\vspace{-6pt}
  \caption{\textit{Learning deformations with SMPL-X vertex guidance on SDFs and point sets}. The usage of the introduced SMPL-X guidance regime benefits both SDF baseline \cite{palafox2021npms} and our reconstructions in the case of the challenging ReSynth scans \cite{POP:ICCV:2021}, with our method providing better reconstruction in the areas where no supervision is available. Please pay attention to the reconstruction of the subject’s dress.}
\label{fig:fig5}
\end{figure}

We consider both unsupervised and weakly supervised scenarios for learning deformations. In the case of ReSynth, this supervision comes in the form of correspondence between minimally clothed human bodies in the target and source scans. For all the pairs, we can also obtain guidance supervision by running Lepard \cite{lepard2021} keypoint matcher; however, the keypoint matching approach often fails on the parts such as loose clothing (see appendix), motivating the need for a more robust weakly supervised method.

\paragraph{Results.} The results are presented in Table \ref{table:dynamic} and Figures \ref{fig:fig4},~\ref{fig:fig5}. Without the full supervision of scene flow originally implied by the framework and rarely available in practice, the SDF baseline NPMs \cite{palafox2021npms} fails to find a plausible deformation in case of complex rapid motion, resulting in high reconstruction error (Table \ref{table:dynamic}, r.1, results for DeformingThings4D, ReSynth) and severe visual artifacts (Figures \ref{fig:fig4}\textit{c}, \ref{fig:fig5}\textit{b}). State-of-the-art point cloud registration approach NDP \cite{li2022non} provides the best results among other baselines; however, the constraints baked into the architecture do not allow it fully deform into the desired shape and deal with high-frequency details, leading to higher reconstruction errors compared to our method (\ref{table:dynamic}, r.4) and geometry distortions (Figure \ref{fig:fig4}\textit{d}, mind the animal model paws). We provide more side-by-side comparisons with the method in the appendix.
Finally, the usage of the introduced SMPL-X guidance regime benefits both SDF baseline and our reconstructions (Figure \ref{fig:fig5}, Table \ref{table:dynamic}, r.9-10) in the case of the challenging ReSynth scans, with our method still providing significantly lower reconstruction errors.
 
\subsection{Animating Avatars in Challenging Clothing} 
\label{subsec:avatars}
\begin{table}[!t]
\resizebox{\linewidth}{!}{
\begin{tabular}{c|cccc}
\hline
Method & SCANimate~\cite{saito2021scanimate} & PoP~\cite{POP:ICCV:2021} & SkiRT~\cite{SkiRT:3DV:2022} & Ours \\
\hline
User choice\% & 6.1\% & 2.9 \% & 4.9\% & \textbf{85.8\%} \\
\hline
\end{tabular}
}
\vspace{0.8pt}
\caption{\textit{Perceptual study results}. 
Across all examples, 85.8\% users prefer results from our method over others.} \label{tab:user_study}
\vspace{-12pt}
\end{table}

\begin{figure*}[tb]
\centering
  \includegraphics[width=0.92\linewidth]{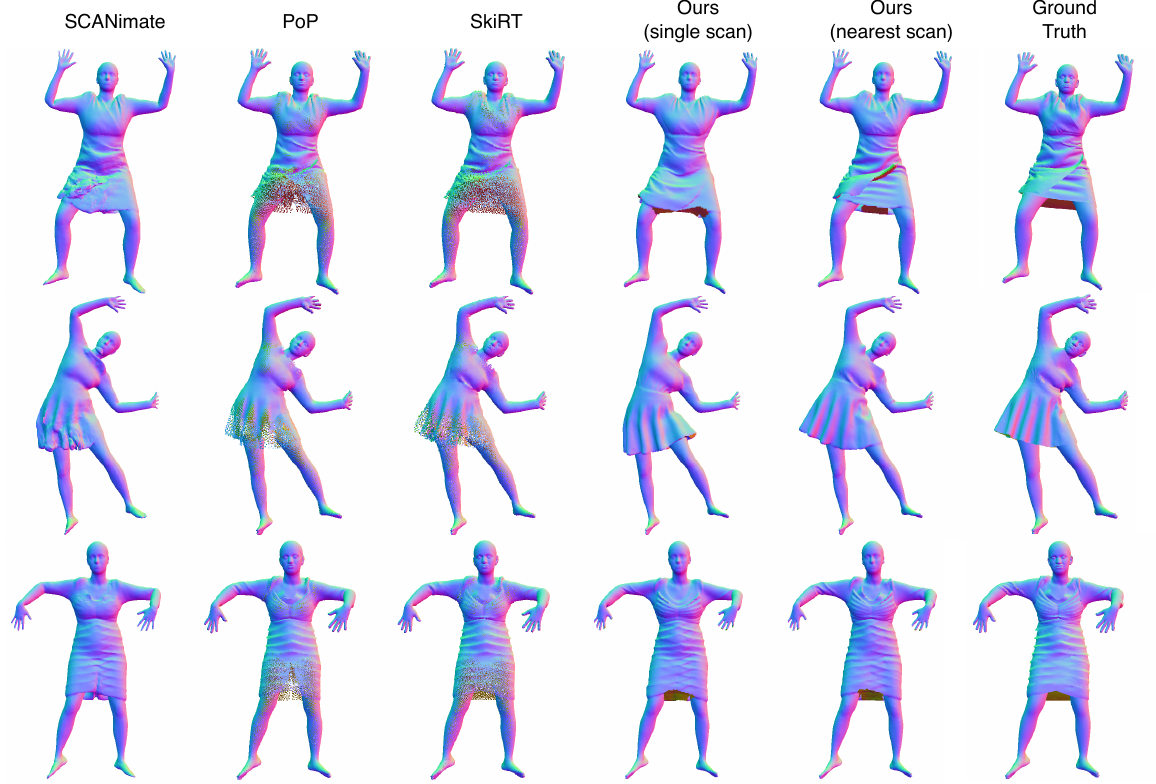}
\vspace{-3pt}
  \caption{\textit{Qualitative comparison of clothed body shape modeling.} SCANimate~\cite{saito2021scanimate} results are rendered from the iso-surface extracted from the SDF, and often erroneously produce trousers-like structure for skirts. PoP~\cite{POP:ICCV:2021} and SkiRT~\cite{SkiRT:3DV:2022} are rendered with surfel-based renderer~\cite{pfister2000surfels,wiles2020synsin} to achieve best visual quality, but still suffer from artifacts due to points being sparse on certain regions. The two variants of our approach both produce vivid wrinkles and a globally coherent clothing shape, with a more salient pose-dependent effect when deforming from the nearest training scan. 
 Best viewed zoomed-in on a color screen.}
 \vspace{-8pt}
\label{fig:comparison_clo}
\end{figure*}

Finally, we showcase the advantages of our approach for modeling shapes of 3D humans in challenging clothing. 

\paragraph{Problem setup and data.} 
On a high level, the task is to generate plausibly-looking clothed body geometry for the test poses given the \{\textit{body pose}, \textit{clothed body shape}\} pairs in the training set. 
We evaluate on the ReSynth~\cite{POP:ICCV:2021} dataset.
It contains 3D humans in simulated clothing, featuring rich geometric details such as wrinkles and folds that vary with changing body poses. 
For each subject, the examples in the training- (637 examples) and test-set (347 examples) have different poses, hence different pose-dependent clothing shapes.
We specifically choose subjects wearing skirts and dresses, as these clothing types have been a long-standing challenge for learning-based 3D clothed human models.

\paragraph{Baselines.} We compare with three state-of-the-art learning-based clothed human models. SCANimate~\cite{saito2021scanimate} models the clothing geometry using SDFs, whereas PoP~\cite{POP:ICCV:2021} and SkiRT~\cite{SkiRT:3DV:2022} deform a point cloud sampled from the unclothed body to represent clothing. Trained in a subject-specific manner, all these methods handle body articulation with Linear Blend Skinning (LBS), and learn a regression from body pose to the clothed body geometry, with an expectation to generalise to novel poses at test time. 

\paragraph{Our approach.} Fundamentally different from the baselines, we sidestep LBS and the pose space generalisation challenge and tackle the task by optimising deformation fields from a canonical scan, using the guided deformation field learning introduced in Sec.~\ref{subsec:training}.
Regarding the choice of canonical frames, we consider two alternatives: (1) using a single scan from the training set as the canonical frame. 
(2) for each test frame, finding its body pose nearest neighbor frame in the training set, and use the scan from that frame as the source of deformation. We provide the details on our pipeline in the appendix Section~\ref{supp:sec:clothing_exp}.

\paragraph{Evaluation metric.} 
Here we conduct a perceptual study among 30 users to characterise the shape quality and visual plausibility of the results. 
We randomly sample subjects and test-set poses, and render predictions from all methods under the same setting. The renderings are placed side-by-side with shuffled ordering across examples. For each example, the users are asked to choose the single result with highest quality.
Further discussions on the metric and details on the user study are provided in the appendix.

%%%%%%%%%%%%%%%%%%%%
\paragraph{Results.} Fig.~\ref{fig:comparison_clo} shows the shape prediction of clothed body from all the methods. While all baselines can generate pose-dependent clothing deformations, they exhibit a variety of artifacts. SCANimate~\cite{saito2021scanimate} suffers from representing thin cloth structures such as the bottom of skirts, a typical challenge for implicit surface representations.
The point-based models PoP~\cite{POP:ICCV:2021} and SkiRT~\cite{SkiRT:3DV:2022} deliver sharp wrinkles where points are dense (e.g. upper body), but lose expressiveness on low-density regions especially for loose-fitting clothing. 
In contrast, our method are free from such artifacts, producing a coherent global shape and sharp local details, essentially animating scans without compromising the original geometry quality.
This verifies the effectiveness of our guided deformation learning on highly non-rigid data.

From the clothed human modeling perspective, due to the reliance on Linear Blend Skinning, the baseline methods often struggle with loose garments and skirts, resulting in e.g. ``splitting'' artifacts on skirts as shown in Fig.~\ref{fig:comparison_clo}.
In contrast, our method offers a new paradigm for clothing modeling, which directly optimises for a smooth deformation field that preserves the continuity of cloth surfaces, side-stepping drawbacks of LBS and the challenges in the pose space generalisation.
In particular, it can produce plausibly-looking clothing surface even under extreme, out-of-distribution poses (Fig.~\ref{fig:fig2}\textit{d}), which, to our knowledge, has not been demonstrated before for learning-based clothed human models. 
More on clothed human modeling paradigms are discussed in the appendix.

These qualitative advantages of our method are also characterised by a clear winning margin in the user study as shown in Table~\ref{tab:user_study}. Overall, 85.8\% users prefer the results produced by our method for higher geometry quality and visual resemblance to the ground truth.

\section{Conclusion}
\label{sec:conclusion}

We have introduced a dynamic point field model that efficiently models non-rigid 3D surfaces by combining explicit point-based graphics with implicit deformation networks. 
Incorporating established constraints like as-isometric-as-possible regularization is made easy by using explicit surface primitives. 
Extensive experiments have shown superiority of our dynamic point field in representational power, learning efficiency and robustness.
We have also demonstrated that our framework offers an advantageous new paradigm for animating clothed humans, surpassing the limitations of linear blend skinning-based methods especially on complex clothing types such as skirts.
Currently our method runs per-frame and does not explicitly model the dependency of clothing shape on body poses. Future work can leverage the advantages of our representation to build e.g. robust, high quality pose-dependent avatar models.

\paragraph{Acknowledgements.} This work was supported by an ETH Z{\"u}rich Postdoctoral Fellowship. Qianli Ma is partially funded by the Max Planck ETH Center for Learning Systems. We sincerely thank Marko Mihajlovic, Yan Zhang, Anpei Chen, Du\v{s}an Svilarkovi\'{c} and Shaofei Wang for the fruitful discussions and manuscript proofreading. 

\newpage
\input{arxiv_appendix}

{\small
\bibliographystyle{ieee_fullname}
\bibliography{references}
}

\end{document}

%% file: macros.tex
\usepackage{iccv}
\usepackage{times}
\usepackage{epsfig}
\usepackage{graphicx}
\usepackage{amsmath}
\usepackage{amssymb}
\usepackage{multirow}
\usepackage[numbers,sort,compress]{natbib}
\usepackage[pagebackref=true,breaklinks=true,letterpaper=true,colorlinks,bookmarks=false]{hyperref}

\DeclareMathOperator*{\argmin}{argmin}
\newcommand{\xb}{\mathbf{x}}

\newcommand*{\affaddr}[1]{#1} 
\newcommand*{\affmark}[1][*]{\textsuperscript{#1}}

%% file: arxiv_appendix.tex
\appendix
{\noindent\Large\textbf{Appendix}}
\counterwithin{figure}{section}
\counterwithin{table}{section}

\section{Extended Details on DPF Method}\label{supp:sec:more_method}
\subsection{Normal Estimation and Transfer}
Complementing main paper Section~\ref{subsec:dpf}, here we elaborate on how to obtain point normals in the warped point clouds. 

We consider three strategies to obtain a new point normal direction $\mathbf{n}_i^{(t)}$. First, in most considered scenarios, the points in canonical space represent vertices of a dense mesh, which are either given as ground truth or obtained by running a surface reconstruction algorithm \cite{kazhdan2013screened}. Hence, after applying the transformation $g_\theta$, we can keep the canonical space connectivity and directly obtain the new normal directions. If no such information is available, or the induced deformations lead to some drastic changes in mesh topology, we can also re-estimate normals using standard approaches \cite{hoppe1992surface}. Please note that all the steps above are fully differentiable, thanks to the modern deep learning 3D frameworks \cite{pytorch3d} and differentiable Poisson solvers \cite{Peng2021SAP}. 

Another way to obtain new normal directions is to directly warp them with the deformation network. We leave for future work an exploration of the correct architecture and parametrisation for such learning-based normal estimation, and provide only a proof-of-concept example of such approach in Figure \ref{fig:sup_fig1}. 

In general, thanks to the introduced isometric loss and a bias of the network towards smooth predictions, the predicted deformations in our experiments allow us to directly reuse the canonical space mesh topology; this strategy is implied, if not stated otherwise.

\subsection{Optimisation Details}

In most cases, we optimise the deformation field for 2000 steps, sampling $10^4$ random points for computing $\mathcal{L}_{CD}$ and $\mathcal{L}_{iso}$. We use Adam optimiser \cite{adam} with an initial learning rate of $10^{-4}$, decreasing the rate by $10^{-1}$ after each $200$ steps of no improvement. 

If a well-defined ground truth canonical surface is given (Sections \ref{subsec:deformation_fields_sdf}-\ref{subsec:avatars} of the experiments), we omit the optimisation of point locations in the canonical space and optimise only the deformation field parameters w.r.t. $\mathcal{L}_{CD}, \mathcal{L}_{iso}, \mathcal{L}_{V}$.

\begin{figure}[t]
\begin{center}
  \includegraphics[trim={0.5cm 0.0cm 0.0cm 0cm},clip,width=1.0\linewidth]{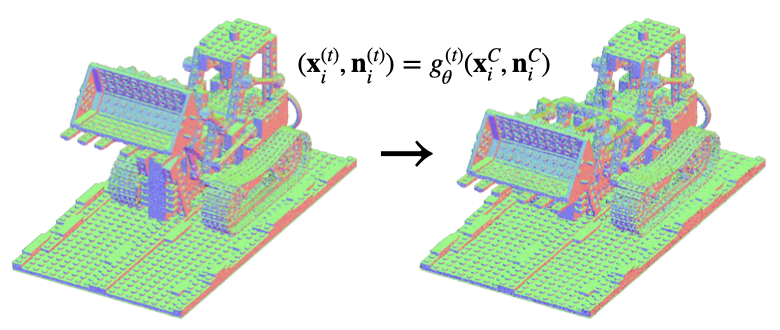}
\end{center}
\caption{\textit{Joint optimisation of the canonical space cloud and deformation network}. In this illustrative example, we showcase the simultaneous optimisation of the canonical cloud $X^{\mathcal{C}}$ and the deformation network $g_\theta$. Apart from new point locations, the deformation network also predicts updated normal directions for each point.  Please see the supplemental video for the animation of the sequence (title slide).} 
  \vspace{-0.3cm}
\label{fig:sup_fig1}
\end{figure}

\begin{figure}[t]
\begin{center}
  \includegraphics[trim={0.0cm 0.0cm 0.0cm 0.0cm},clip,width=1.0\linewidth]{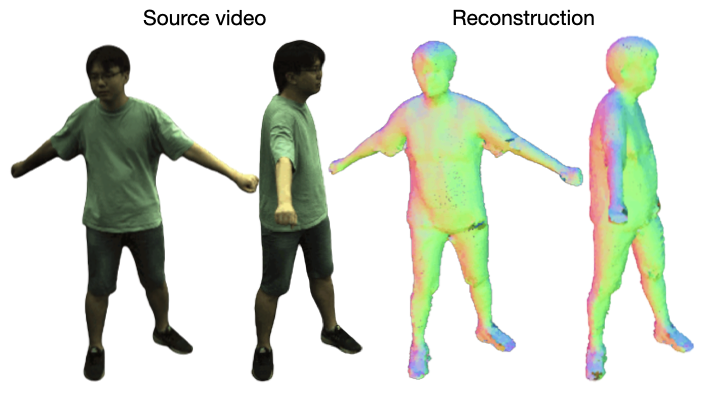}
\end{center}
\vspace{-0.3cm}
  \caption{\textit{Dynamic point field reconstruction from multi-view video streams}. As an avenue for future work, we consider the reconstruction of dynamic surfaces from multi-view videos. Early results on the sequence from \cite{peng2021neural}.} 
  \vspace{-0.3cm}
\label{fig:sup_fig2}
\end{figure}

However, the possibility to optimise canonical space together with the deformation framework and to utilise rendering-based losses (e.g. $\mathcal{L}_{n_I}$) is essential for future work, when the introduced deformation paradigm will be combined with efficient photorealistic point-based renderers \cite{zhang2022differentiable,xu2022point}. We showcase feasibility of such optimisation in Figures \ref{fig:sup_fig1}-\ref{fig:sup_fig2}. 

\section{Extended Experiments, Results, Discussions}

\subsection{Hyperparameter Selection, Implementation}

For all the baselines considered in the paper, we aim to find the optimal parameters for optimisation and rendering.  E.g., we experiment with the number of elements in the deformation pyramid \cite{li2022non} to achieve best results, find the best hyperparameters for the marching cube algorithm when comparing to SDF-based methods, etc. 

We use original implementations of the NGLOD \cite{takikawa2021nglod}, NPMs \cite{palafox2021npms}, NDP \cite{li2022non} and Siren \cite{sitzmann2019siren} frameworks \cite{nglod_code,npms_code,ndp_code,siren_code}. We use the torch-NGP framework \cite{torch-ngp} for the instant NGP \cite{mueller2022instant} baseline since it allows easy modification, and \cite{ndp_code} for the Nerfies \cite{park2021nerfies} and neural scene flow prior \cite{li2021neural} baselines. For the avatar animation baselines \cite{POP:ICCV:2021,saito2021scanimate,SkiRT:3DV:2022}, we use the results kindly provided by the authors.

\subsection{Point-based Surface Reconstruction}

\noindent\textbf{Naive point sampling.} Apart from the baselines already addressed in Section~\ref{subsec:static_experiments}, a straightforward candidate for comparison is a point cloud acquired directly by sampling points and respective normals from the ground truth surface, with no training involved. This simple method of storing surfaces gives good Chamfer distance reconstruction results by its very nature since the surface is represented by its samples. However, it suffers from visual artifacts when rendering due to the overlapping points and noisy normal directions (see Figure~\ref{fig:fig3}\textit{a}, and slide \textcolor{red}{4} of the supplemental video). Optimizing these initial point sets with our method not only produces better normal directions but also further improves Chamfer distance metrics since the optimization drives the points to be better distributed along the surface. This is especially noticeable in the case of a limited sample size (Table \ref{table:static}).

\begin{figure*}[t!]
\centering
\includegraphics[trim={0.0cm 0.0cm 0.0cm 0.0cm},clip,width=\textwidth]{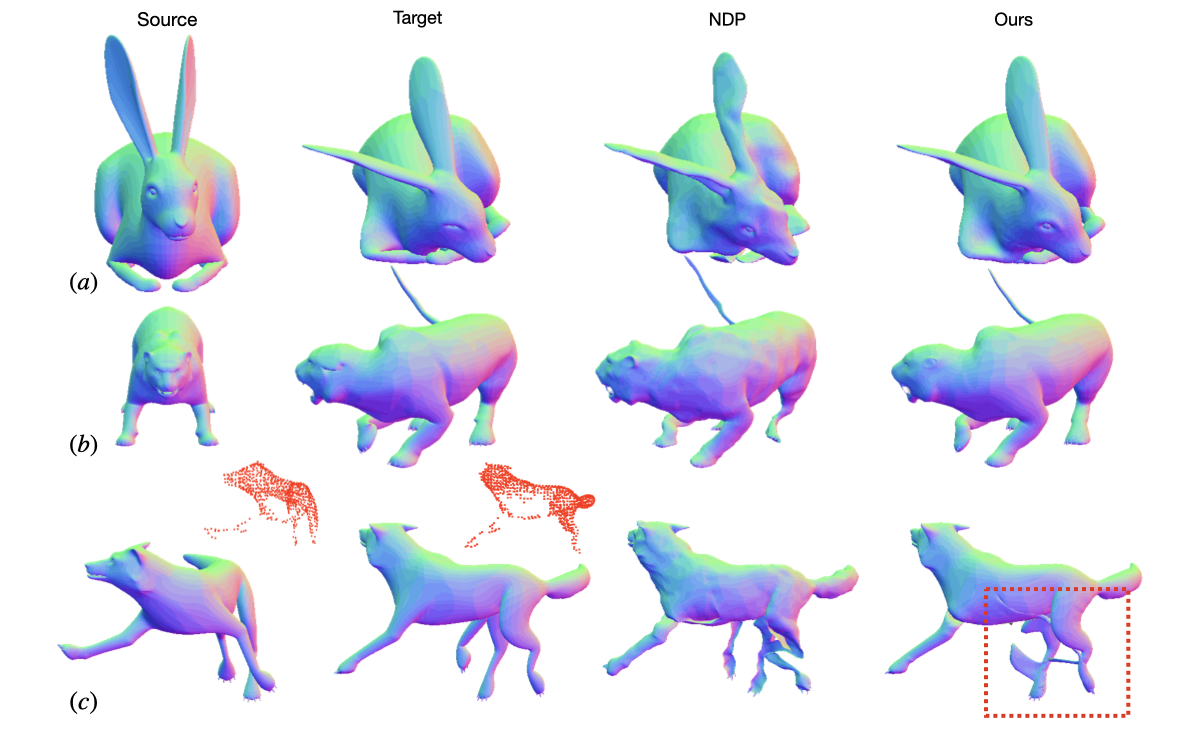}
\vspace{-0.4cm}
\caption{\textit{Additional results on non-rigid surface registration.} (a-b): comparison to the NDP \cite{li2022non} method in the case of Lepard \cite{lepard2021} keypoint supervision. (c): failure case of our method in the situation of the highly articulated pose with missing keypoints. Matched points are visualised in red, pay attention to missing limbs. } 
\label{fig:sup_fig3}
\end{figure*}

\noindent\textbf{Other techniques.} The proposed point optimization scheme can also be considered a hybrid of the differentiable surface splatting technique \cite{Yifan:DSS:2019} and shapes-as-points approach \cite{Peng2021SAP}. Compared to DSS, we additionally utilize the Chamfer distance loss and a more efficient point renderer \cite{wiles2020synsin}, which results in better reconstruction. Compared to SAP, our method has lower training time memory requirements since no differentiable Poisson solver needs to be ran on the grid for each pass. However, this technique can be complementary to our basic pipeline in case of the reconstruction from unoriented, noisy point clouds, as well as when a water-tight mesh reconstruction is required. The actual goal of this experiment is not to claim the ultimate advantage of any particular point optimization scheme, but rather show their appealing features compared to implicit methods.

\begin{figure}
\begin{center}
  \includegraphics[trim={0.0cm 0.0cm 0.0cm 0.0cm},clip,width=1.0\linewidth]{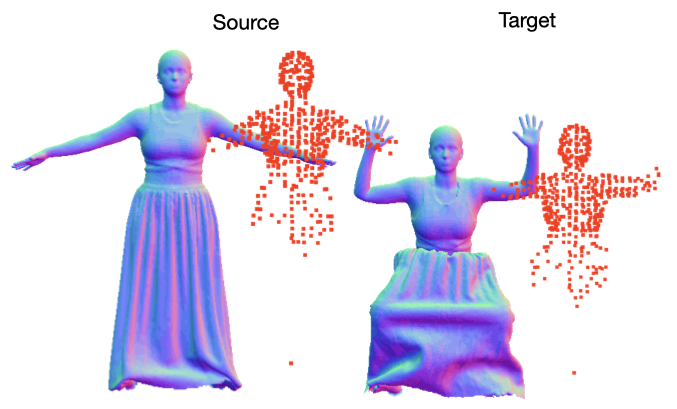}
\end{center}
\vspace{-0.3cm}
  \caption{\textit{Detected Lepard \cite{lepard2021} keypoints on the Resynth scan pair}. The off-the-shelf landmark detection method fails in the case of highly articulated humans and loose clothing, motivating the need for better deformation supervision.} 
  \vspace{-0.3cm}
\label{fig:sup_fig4}
\end{figure}

\begin{center}
\begin{table}
\footnotesize
\centering
\resizebox{\columnwidth}{!}{
\begin{tabular}{l|c|c|c|c}
Method                                                                & $\mathcal{L}_{CD}$ $\downarrow$ & $\mathcal{L}_{n}$ $\downarrow$  & Size (Mb) & Params ($10^6$)  \\
\hline 
 Sampled points ($10^4$)   & 5.522                & 0.680                 & 0.24   & 0.06                       \\
  Sampled points ($10^5$) & 1.101                & 0.554                & 2.4    & 0.6                        \\   Sampled points ($3\cdot10^5$) & 0.536                & 0.473                & 7.2    & 1.8                        \\
  Sampled points ($10^6$)  & 0.274                & 0.378                & 24      & 6                          \\
  Ours ($10^4$ points)     & 3.362                & 0.626                & 0.24    & 0.06                       \\
  Ours ($10^5$) & 0.765                & 0.468                & 2.4     & 0.6                        \\
  Ours ($3\cdot10^5$) & 0.419                & 0.411                & 7.2     & 1.8                        \\
  Ours ($10^6$) & \textbf{0.246}       & \textbf{0.322}       & 24      & 6        \\
  \hline 
\end{tabular}
}
\vspace{0.8pt}
  \caption{\textit{Static surface reconstruction: optimised point sets vs random samples.}}
  \label{table:static}
\end{table}
\end{center}

%%%%%%%%%%%%%%%%%%
\subsection{Experiment: Learning Deformations}\label{supp:sec:learning_deformations}

\paragraph{Additional results.} Augmenting the discussion in Section~\ref{subsec:deformation_fields_sdf} of the main manuscript, we provide additional comparisons with the best-performing non-rigid registration  method \cite{li2022non} on the DeformingThings4D dataset in Figure \ref{fig:sup_fig3}, as well as the failure case of our method (c).

\paragraph{SMPL-X guidance vs keypoint matching.} To qualitatively justify the advantages of the SMPL-X based deformation learning guidance (Section~\ref{subsec:deformation_fields_sdf}), we visualise and discuss the results of the 3D keypoint matching algorithm \cite{lepard2021} in Figure \ref{fig:sup_fig4}. 

%%%%%%%%%%%%%%%%%%
\subsection{Experiment: Avatar Animation}\label{supp:sec:clothing_exp}
Here we provide extended details on the experiments in the main manuscript Section~\ref{subsec:avatars}. 

We conduct the experiment on the ReSynth~\cite{POP:ICCV:2021} dataset. Each data frame $t$ contains a dense point cloud $\{\mathbf{x}_i^{(t)}\}_{i=1,\cdots,N}$ ($N\approx2\times10^5$ points) of a posed, clothed person , and its corresponding underlying minimally-clothed body mesh vertices $\{\mathbf{v}^{(t)}\}_{i=1,\cdots,N_v}$. 
The mimnimal body meshes have a consistent topology across all frames, but the clothed body point clouds do not have temporal correspondence.
Given a set of such data pairs, we aim to produce clothed body for unseen test poses. 

\paragraph{Implementation details.}
We adapt the guided deformation field learning in the main manuscript Section~\ref{subsec:training}. On a high level, we optimise a per-frame MLP to model the deformation field between minimal bodies of the canonical- and the target frame, and apply the optimised MLP to deform clothed body surface points. 

Specifically, to predict a test frame $t$, we choose a canonical data frame $\mathcal{C}$ from the training set, and optimise a deformation field parameterized by an MLP $g_\theta^{(t)}$, such that the minimally-clothed body mesh vertices in the canonical frame $\{\mathbf{v}_i^{(\mathcal{C})}\}_{1,\cdots,N_v}$ matches that of the target frame $\{\mathbf{v}_i^{(t)}\}_{1,\cdots,N_v}$ after transformed by $g_\theta^{(t)}$:

\begin{align}\label{eq:supp_clo_objecctive}
\begin{split}
    \theta^* = \argmin_{\theta} \sum_{i=1}^{N_v} & \lambda_V||\mathbf{v}^{(t)}_i - (\mathbf{v}^{(\mathcal{C})}_i + g_\theta(\mathbf{v}^{(\mathcal{C})}_i)||_1\\
    & + \lambda_{iso} \mathcal{L}_{iso} .
\end{split}
\end{align}
The optimisation is performed using the Adam optimiser with a learning rate of $1\times10^{-4}$ and hyperparameters $\lambda_{iso}=10^3, \lambda_{V}=10^4$ for 2000 steps.

Once optimised, $g_\theta^{(t)}$ is used to transform arbitrary points on the canonical clothed body surface into the target pose. In practice, instead of deforming all points in the high-resolution source point cloud, we first perform Poisson Surface Reconstruction (PSR)~\cite{kazhdan2013screened} to obtain a mesh and deform the mesh vertices. 
In this way, we can transfer the mesh connectivities derived from PSR to the deformed point set and directly obtain the points' normals without the need to predict or re-estimate them, as discussed in Section~\ref{supp:sec:more_method}.
Although the deformation field $g_\theta^{(t)}$ is optimised for the minimally-clothed body, the smooth inductive bias of the coordinate-based MLP results in a coherent clothing shape after deformation. 

\paragraph{Choice of canonical frame.}
We investigate two strategies of choosing the canonical frame. (1) \textit{``Single canonical frame''}: we simply choose the first frame (in an ``A''-pose) from the ReSynth training set and use that as the source of deformation for all test poses. 
Empirically we find that the results from this approach are temporally coherent, as shown on the slide \textcolor{red}{16} of the supplemental video. 
It is also worth noting that this setting is essentially a solution to the ``single scan animation'' challenge~\cite{POP:ICCV:2021}, i.e. animating a given single scan to different novel poses.
(2) \textit{``Nearest canonical frame''}: for each test pose, we find its nearest pose (measured by the vertex-to-vertex distance of the underlying body mesh) in the training set, and use that frame as its canonical frame. This approach does not guarantee temporal consistency due to the discontinuity nature of nearest neighbor search, but the produced clothing geometry varies more obviously with the body pose.

\paragraph{Discussions on the evaluation metric.}
In the main paper we compared with baseline methods using a perceptual study instead of the Chamfer distance and normal consistency metrics as used in recent papers~\cite{SkiRT:3DV:2022,POP:ICCV:2021,saito2021scanimate}. Here we discuss the reason.

\begin{figure}
\includegraphics[width=\linewidth]{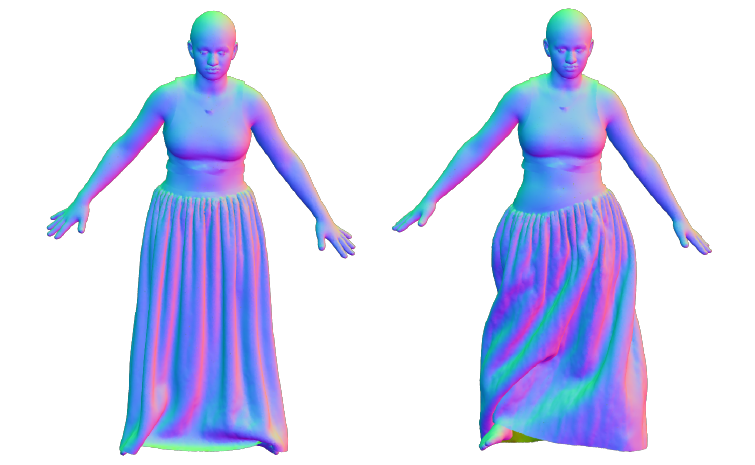}
\caption{\textit{Stochastic clothing shape.} Given very similar body poses, the clothing shape state can largely differ, while both being valid. Example taken from the ReSynth~\cite{POP:ICCV:2021} dataset, from the beginning and the end of the ``hips'' motion sequence, respectively. The body dynamics (motion history) has influenced the clothing shape despite the similarity in pose. The chamfer distance and normal consistency error between these two frames are $30.0\times10^{-4}m^2$ and $3.5\times10^{-1}$, respectively.}\label{fig:supp_clo_stochasticity}
\end{figure}

Previous works~\cite{POP:ICCV:2021,SkiRT:3DV:2022,saito2021scanimate} evaluate clothing shape prediction accuracy by computing Chamfer and surface normal distances to the ground truth. This implicitly assumes a one-to-one mapping from body pose to the clothing shape but in reality,
the clothing shape is not solely dependent on pose; it is also influenced by other factors such as the motion speed and history. Consequently, for a given pose, multiple clothing shapes can be plausible, as discussed in ~\cite{CAPE:CVPR:20,bagautdinov2021driving}. An illustration is shown in Figure~\ref{fig:supp_clo_stochasticity}.

Since the experiment in Section~\ref{subsec:avatars} is primarily purposed to demonstrate the representational power of our method and its applicability to human modeling, we resort to a perceptual study to characterize its quality, instead of adopting the pose-only regression metrics as discussed above.

Nevertheless, here we also provide the Chamfer distance and normal consistency errors for a reference in Table~\ref{tab:supp_clo_chamfer}. 
As our method deforms a source (canonical) cloud, the target state it produces, while plausibly-looking, may not conform that of the ground truth, hence the higher errors in the table.
While pose-shape regression is not the focus of this work, we believe that combining it with our representation can lead to comparable to lower errors under these metrics while achieving higher visual quality as we show in the paper. We leave this for future work.

\begin{table}[!h]
\centering
\resizebox{\linewidth}{!}{
\begin{tabular}{rllllllllll}
\hline
\multirow{2}{*}{Method} & \multicolumn{2}{c}{anna-001}                     & \multicolumn{2}{c}{beatrice-025}                 & \multicolumn{2}{c}{christine-027}                & \multicolumn{2}{c}{janett-025}                   & \multicolumn{2}{c}{felice-004}                   \\ \cline{2-11} 
                        & \multicolumn{1}{c}{CD} & \multicolumn{1}{c}{NML} & \multicolumn{1}{c}{CD} & \multicolumn{1}{c}{NML} & \multicolumn{1}{c}{CD} & \multicolumn{1}{c}{NML} & \multicolumn{1}{c}{CD} & \multicolumn{1}{c}{NML} & \multicolumn{1}{c}{CD} & \multicolumn{1}{c}{NML} \\ \hline
SCANimate~\cite{saito2021scanimate}               & 1.34                   & 1.35                    & 0.74                   & 1.33                    & 3.21                   & 1.66                    & 2.81                   & 1.59                    & 20.79                  & 2.94                     \\
PoP~\cite{POP:ICCV:2021}                     & 0.62                   & 0.82                    & 0.34                   & 0.75                    & 1.72                   & 0.97                    & 1.24                   & 0.89                    & 7.34                   & 1.24                     \\
SkiRT~\cite{SkiRT:3DV:2022}                   & 0.58                   & 0.81                    & 0.31                   & 0.77                    & 1.54                   & 0.99                    & 1.10                   & 0.82                    & 6.45                   & 1.25                     \\
Ours (single)      & 1.27                   & 0.99                    & 0.68                   & 0.95                    & 4.40                   & 1.39                    & 3.12                   & 1.36                    & 14.45                  & 2.56                     \\
Ours (nearest)     & 0.96                   & 0.96                    & 0.46                   & 0.99                    & 2.88                   & 1.24                    & 2.51                   & 1.19                    & 16.07                  & 2.50                  \\
\hline
\end{tabular}
}
\vspace{4pt}
\caption{Errors on pose-only regression metrics. Chamfer distance in $\times10^{-4}m^2$; normal consistency in $\times10^{-1}$. ``Ours (single)'': using a single frame as canonical frame; ``Ours (nearest)'': using nearest training scan per frame as the canonical frame (see Section~\ref{supp:sec:clothing_exp}, \textit{``Choice of canonical frame''}).}
\label{tab:supp_clo_chamfer}
\end{table}

\begin{figure*}
\centering
\includegraphics[width=0.95\linewidth]{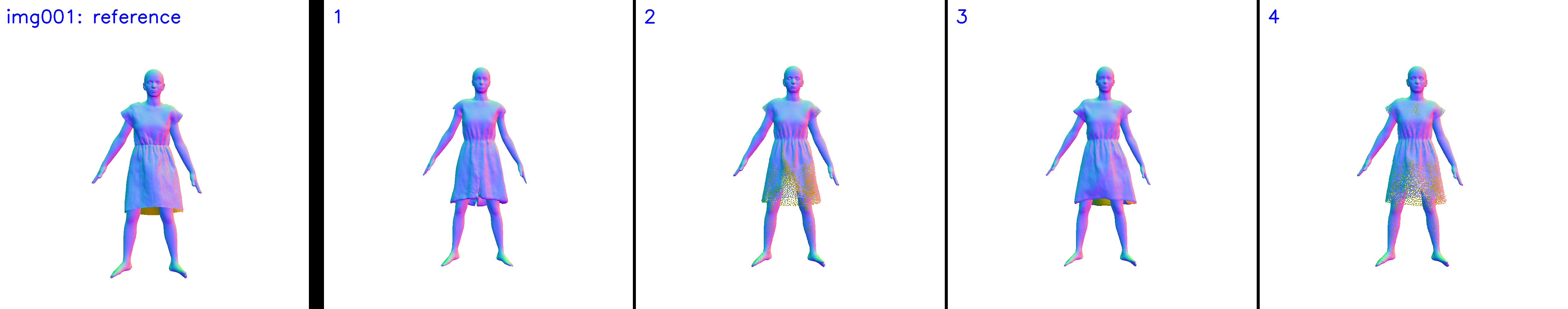}
\caption{\textit{Example of perceptual study image.} The ground truth rendering is always presented on the leftmost position, and the ordering of results from different methods is shuffled per example.}\label{fig:supp_perceptual_study}
\end{figure*}

\begin{figure}
    \centering
    \includegraphics[width=\linewidth]{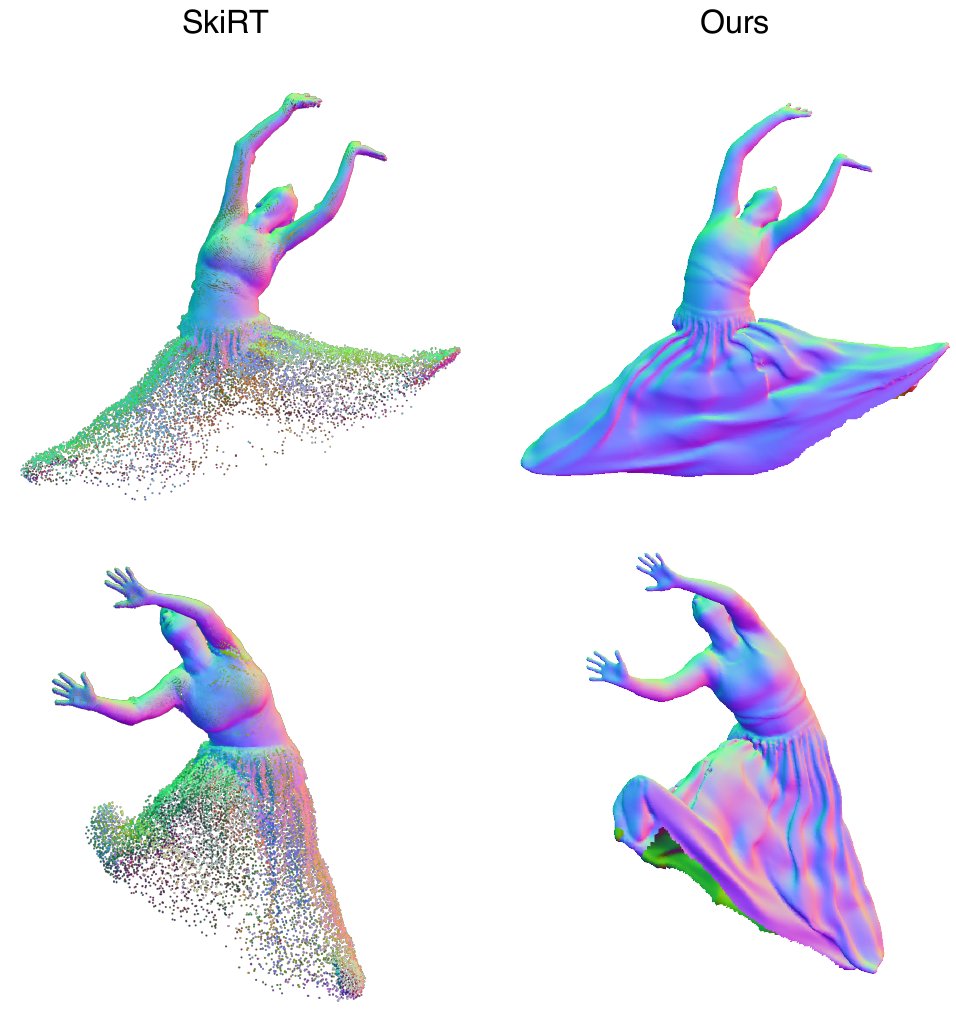}
    \caption{Qualitative comparison against the SoTA point-based clothed human model~\cite{SkiRT:3DV:2022} under extreme poses.}
    \label{fig:supp_clo_extreme_pose}
\end{figure}

\paragraph{Details on perceptual study.}
We choose the optimal renderering parameters for each method. For point-based baselines~\cite{POP:ICCV:2021,SkiRT:3DV:2022}, we use PyTorch3D point renderer with a point radius of 0.007 to minimize the visual artifact caused by the gaps between points. For SCANimate, we first extract a mesh from the generated implicit surface and render with the PyTorch3D mesh renderer. For our method we use the results generated by the \textit{Nearest canonical frame} scheme, perform PSR (see paragraph \textit{``Implementation details''}) and render the deformed mesh. The camera and lighting condition are kept same for all renderings. 

The participants are presented with a set of 32 examples consisting of different subjects and poses, randomly sampled from the ReSynth dataset results. In each example, the ground truth rendering is shown on the left, and the results generated by different methods are presented in random order on the right, see Figure~\ref{fig:supp_perceptual_study}. For each example, the participants are asked to select their most preferred, single result based on the following two criteria: (1) overall visual quality considering the realism of clothing shape, wrinkle details and level of artifacts; (2) resemblance to the ground truth. 
The statistics in the main paper Table~\ref{tab:user_study} are computed by dividing the number of votes each method receives with the total number of votes (excluding $0.3\%$ empty votes).

\paragraph{Result: extreme poses.} Figure~\ref{fig:supp_clo_extreme_pose} shows further qualitative results on predicting the clothing shape under extreme, out-of-distribution poses, in comparison to a recent point-based clothed human model, SkiRT~\cite{SkiRT:3DV:2022}.
Although SkiRT is designed to address the ``split''-like artifacts for skirts and dresses by learning an LBS field for the garment, it inevitably reaches its limit under these poses: the points between the legs become very sparse.
In contrast, our method well retains the completedness and smoothness of the cloth, showing a higher robustness for these garment types under extreme poses.

\section{Limitations and Future Work}
\label{supp:sec:limitation}

Our deformation optimisation pipeline still struggles to find a plausible deformation when large deformations are present and no guidance is available (Figure \ref{fig:sup_fig3}\textit{c}). This is especially pronounced in the case when the scene undergoes significant topological changes. One way to improve the deformation learning in the case of animals is to use estimators of the corresponding body joints \cite{Mathisetal2018,nath2019using} and 3D models \cite{zuffi20173d,Zuffi_2019_ICCV}.

Second, as discussed in the main paper, currently our method requires optimising a small network for each frame when dealing with unseen target scans or poses, which takes approximately 30\textit{s}/frame with a NVIDIA RTX6000 GPU for the ReSynth data. This makes our method, in its current form, unsuitable for real-time applications. Potential solutions here include meta-learning approaches to learning deformations networks \cite{sitzmann2020metasdf,wang2021metaavatar}.

Finally, when applied to modeling humans, the guided deformation optimisation does not explicitly learn the dependence of the clothing geometry on factors such as body pose and acceleration, which is an open research topic per se. 
Nevertheless, our experiments have demonstrated the capability of our approach in modeling highly non-rigid, articulated shape such as dynamic humans. Future work can combine our representation with domain-specific techniques to create high-quality real-time avatar models with explicit control on semantic and physical parameters.